\documentclass[review]{elsarticle}

\usepackage{lineno,hyperref}
\modulolinenumbers[5]

\usepackage{amsmath,amsthm,amssymb,epsfig,bm,amsmath}
 \usepackage{graphicx}
\usepackage{float}
\usepackage{csquotes}
\usepackage{amsmath}
\usepackage{mhchem}
\usepackage{xparse}
\usepackage{adjustbox}
\usepackage{MnSymbol}% http://ctan.org/pkg/mnsymbol
\usepackage{mathdots}% http://ctan.org/pkg/mathdots
\usepackage{makecell}
\usepackage{listings}
\usepackage{footnote}
\usepackage[flushleft]{threeparttable}
\usepackage{multirow}
\usepackage{relsize}
\usepackage{lettrine}
\usepackage{courier}
\usepackage{color} %red, green, blue, yellow, cyan, magenta, black, white
\definecolor{mygreen}{RGB}{28,172,0} % color values Red, Green, Blue
\definecolor{mylilas}{RGB}{170,55,241}
\usepackage{amsmath,amsthm,amssymb,amsfonts} % this for /equation* (without numbering)
\usepackage{graphicx} % for figures
\usepackage[margin=1in]{geometry} % changing the margins
\usepackage[yyyymmdd]{datetime}
\usepackage{algorithm}
\usepackage[noend]{algpseudocode}
\usepackage{lipsum}
\usepackage{setspace}
\usepackage{siunitx}
\usepackage{pdfpages}
\usepackage{tikz}
\usetikzlibrary{positioning}
\usepackage{booktabs,caption}
\usepackage[]{hyperref}
\hypersetup{
 colorlinks  = true, %Colours links instead of ugly boxes
 urlcolor   = blue, %Colour for external hyperlinks
 linkcolor  = black, %Colour of internal links
 citecolor  = blue %Colour of citations
}
\usepackage[numbers]{natbib}
\usepackage[utf8]{inputenc}

\usepackage{listings}
\usepackage{color}
\definecolor{codegreen}{rgb}{0,0.6,0}
\definecolor{codegray}{rgb}{0.5,0.5,0.5}
\definecolor{codepurple}{rgb}{0.58,0,0.82}
\definecolor{backcolour}{rgb}{0.95,0.95,0.92}
\lstdefinestyle{mystyle}{
  backgroundcolor=\color{backcolour},  
  commentstyle=\color{codegreen},
  keywordstyle=\color{magenta},
  numberstyle=\tiny\color{codegray},
  stringstyle=\color{codepurple},
  basicstyle=\footnotesize,
  breakatwhitespace=false,     
  breaklines=true,         
  captionpos=b,          
  keepspaces=true,         
  numbers=left,          
  numbersep=5pt,         
  showspaces=false,        
  showstringspaces=false,
  showtabs=false,         
  tabsize=2,
  escapeinside={<@}{@>},
}
 
\usepackage{booktabs} % For \toprule, \midrule and \bottomrule
\usepackage{siunitx} % Formats the units and values

\usepackage{mathtools} % for arrows
\usepackage{gensymb} % To write degree C 
\usepackage{mhchem} % To write chemical symbols
\usepackage{fixltx2e} % To write subscripts and superscripts 
\usepackage{bbm}

\usepackage{multirow}

\usepackage{fancyhdr} % customize the footer and header
\theoremstyle{definition}

\theoremstyle{definition}

\theoremstyle{remark}

\usepackage{soul} %To highlight the text 
\soulregister\cite7
\soulregister\ref7
\soulregister\pageref7
\usepackage{bm}

\usepackage{framed} % Framing content

\usepackage{multicol} % Multiple columns environment
\usepackage{booktabs}

\usepackage{tabularx}
\usepackage{array}
\usepackage{nomencl} % Nomenclature package
\usepackage{tikz}
\makenomenclature
\usepackage[resetlabels,labeled]{multibib}

\renewcommand*\nompreamble{\begin{multicols}{2}}

\renewcommand*\nompostamble{\end{multicols}}

\usepackage{amssymb}% http://ctan.org/pkg/amssymb
\usepackage{pifont}% http://ctan.org/pkg/pifont
\definecolor{light-gray}{gray}{0.95}

\journal{Journal to be decided}

\begin{document}

%\includepdf{nn_title.pdf}

\begin{frontmatter}

\title{\large Generative Reconstruction of Full-Field Sensor States for Safety-Critical Digital Twins}

%Generative Reconstruction of Full-Field Sensor States for Safety-Critical Digital Twins

\author{Umme Mahbuba Nabila$^{a}$, Linyu Lin$^{b}$, Majdi I.~Radaideh$^{a,c,*}$}

\cortext[mycorrespondingauthor]{Corresponding Authors: Umme Nabila (unabila@umich.edu), Majdi I.~Radaideh (radaideh@umich.edu)}
\address{$^{a}$Department of Nuclear Engineering and Radiological Sciences, University of Michigan, Ann Arbor, MI 48109, United States}

\address{$^{b}$Nuclear Science \& Technology Division, Idaho National Laboratory, Idaho Falls, ID 83415, United States}

\address{$^{c}$Department of Computer Science and Engineering, University of Michigan, Ann Arbor, MI 48109, United States}

\begin{abstract}
\small
Accurate reconstruction of incomplete sensor fields is critical for state awareness in safety-critical digital twins (DTs). Thermocouples in high-temperature and radiation environments may drift, fail, or become intermittently unavailable, causing partial observability. This study develops a sparse-sensor reconstruction framework for Oregon State University's High Temperature Test Facility (HTTF) and evaluates six methods under intra- and inter-experiment protocols: three conditional generative models (i.e., a diffusion model [DM], variational autoencoder [VAE], and generative adversarial network [GAN]), two deterministic recurrent neural-network models, and ridge regression. For intra-experiment reconstruction within the PG{-}29 depressurized conduction cooldown (DCC) test, the models combine graph-based spatial encoding with gated recurrent units. The graph neural network (GNN)--gated recurrent unit (GRU) VAE achieves the lowest error, with a mean absolute error (MAE) of $11.73,^\circ$C, a root mean square error (RMSE) of $20.54,^\circ$C, and an $R^2$ of 0.98. The deterministic GNN--GRU and plain GRU obtain RMSE values of $25.26,^\circ$C and $35.22,^\circ$C, respectively, demonstrating the benefit of spatial modeling. All three generative models reduce RMSE by more than half compared with ridge regression. The DM maintains an $R^2$ of 0.94 when known sensors decrease from 91 to 45. For inter-experiment reconstruction, models trained on PG{-}27 pressurized conduction cooldown (PCC) are evaluated on PG{-}29 using section-context conditioning. The section-context DM achieves the best transfer performance, with an RMSE of $42.84,^\circ$C and an $R^2$ of 0.91. These results demonstrate the value of spatiotemporal generative modeling for missing-sensor reconstruction in safety-critical DT applications.

\end{abstract}

\begin{keyword}
Digital twins, Generative AI, High Temperature Test Facility, Diffusion models, Variational autoencoder, Sparse-sensor reconstruction 
\end{keyword}

\end{frontmatter}

%\linenumbers

% \pagenumbering{gobble}
\setstretch{1.3}

\section{Introduction}
\label{sec:intro}

Safety-critical engineered systems operate under conditions in which incomplete or inaccurate knowledge of the physical state can directly affect monitoring, diagnosis, control, and risk-informed decision-making. Such systems are often characterized by tightly coupled physical processes, transient operating conditions, and distributed instrumentation that provides only a partial representation of the underlying state. Nuclear thermal-hydraulic systems provide a representative and particularly demanding example~\cite{todreas2021nuclear,pope2009thermal}. In nuclear reactors and transient experimental facilities, system behavior arises from the interaction of heat conduction, forced and natural convection, thermal radiation, gas-phase mixing and stratification, and flow redistribution through geometrically complex regions. During safety-relevant transients---such as loss of forced flow, depressurization, or breach of a coolant boundary---these mechanisms interact nonlinearly across multiple spatial and temporal scales, producing nonuniform and history-dependent temperature and flow fields that are difficult to characterize from limited observations. Distributed measurements from thermocouples, pressure transducers, flow instruments, and other sensors therefore provide a critical link between the evolving physical system and the monitoring, diagnostic, and decision-support functions used to assess it. As in other safety-critical applications, the reliability of these functions depends not only on the predictive capability of the underlying models, but also on the accuracy, availability, and spatial coverage of the measurements used to represent the system state.

Data-driven modeling increasingly complements physics-based simulation for extracting state information from complex engineered systems. Within nuclear and advanced-energy applications, machine learning has been used to analyze high-dimensional simulation responses and uncertainty~\cite{radaideh2020analyzing}, construct probabilistic surrogates of computationally expensive simulations~\cite{radaideh2020surrogate}, combine simulation and data with uncertainty quantification~\cite{radaideh2019combining}, and calibrate models against experimental measurements~\cite{radaideh2022model}. Multifidelity surrogate modeling has also been applied to transient phenomena such as the onset of natural circulation following depressurized loss of forced cooling in high-temperature gas reactors~\cite{eaheart2026multifidelity}. More broadly, recent reliability research has emphasized learning from partially observed systems by combining limited measurements, physical information, and uncertainty quantification~\cite{xu2025physics}. These developments extend machine learning beyond computational acceleration toward state estimation; for safety-critical applications, however, such models must remain informative when measurements become sparse, degraded, or unavailable.

Digital twins (DTs) provide one framework for integrating sensing, physics-based simulation, and data-driven models into state-estimation and decision-support capabilities~\cite{kropaczek2023digital}. In nuclear applications, DT concepts have been investigated for remote and autonomous monitoring~\cite{hergesheimer2026developing}, state synchronization and online calibration~\cite{xue2025preliminary}, probabilistic representation of transient thermal-hydraulic experiments~\cite{burnett2025variational}, and uncertainty quantification through active learning~\cite{nabila2026active}. Systematic model-development frameworks~\cite{myers2025pymaise} and interpretable machine-learning approaches~\cite{panczyk2025opening} further illustrate the growing emphasis on dependable use of data-driven models in advanced-energy applications. Across these implementations, a common requirement is access to sufficiently informative and trustworthy measurements of the evolving physical system.

That requirement cannot generally be guaranteed in safety-critical systems. Instrumentation may experience degradation, calibration drift, intermittent availability, or outright failure, while physical and economic constraints can limit redundancy. These challenges are particularly pronounced in nuclear environments because of elevated temperatures, radiation-induced material degradation, and accessibility constraints~\cite{hashemian2010aging,scervini2013transmutation}. More generally, data scarcity has been recognized as a reliability challenge for condition monitoring, motivating physics-informed learning approaches that augment limited sensor information~\cite{sanchez2025addressing}. Instrumentation reports from Oregon State University's High Temperature Test Facility (HTTF), for example, document malfunctioning, intermittent, and inconsistent thermocouple channels over its operational history~\cite{woods2019osu}. HTTF therefore provides a representative safety-critical transient system in which maintaining state awareness cannot rely on complete sensor availability.

Data-driven fault and anomaly detection provides one layer of protection against unreliable measurements. Recurrent and convolutional autoencoders have been used to identify anomalies in time-dependent power-electronics signals~\cite{radaideh2022time}, ensemble learning has enabled early detection of developing equipment faults~\cite{radaideh2023early}, and conditional generative models have been investigated for fault prediction in accelerator power systems~\cite{alanazi2023multi}. Recent reliability studies further recognize that measurements from multiple sensors can carry different levels of confidence, motivating adaptive sensor fusion with uncertainty quantification~\cite{yu2026novel}. Fault identification and sensor-quality assessment, however, do not recover the information lost when a measurement is rejected or becomes unavailable. State reconstruction provides a complementary capability: detection determines which information can be trusted, whereas reconstruction seeks to preserve situational awareness from the trustworthy measurements that remain.

Previous work has explored DT frameworks combining forward prediction, probabilistic calibration, and data-driven reconstruction for transient state estimation~\cite{burnett2026variational}. The present study does not seek to develop or evaluate a particular DT architecture; instead, it isolates the underlying reconstruction problem as a general challenge for safety-critical systems: estimating spatially distributed states when only a subset of sensor measurements remains available and trustworthy. Conditional generative modeling is investigated as one methodology for this problem and compared with deterministic learning and regression approaches. This direction is consistent with emerging reliability applications of generative representations for system health monitoring under limited and partially available information~\cite{costa2025few}. Nuclear thermal-hydraulic transient experiments provide the demonstration considered here because their strongly coupled, time-dependent behavior and realistic instrumentation limitations offer a challenging environment for reconstruction under partial observability.

Sparse sensing can arise through two complementary mechanisms. The first is deliberate: complex engineered systems may contain many candidate measurement locations, while physical, economic, computational, and data-throughput constraints make exhaustive instrumentation or assimilation impractical. Highly redundant and correlated measurements can also increase the computational and numerical burden of real-time data assimilation. A smaller but informative subset may therefore be preferable, motivating optimal sensor-placement strategies that preserve reconstruction accuracy while accounting for physical and uncertainty constraints. Within nuclear applications, such approaches have been investigated for TRISO fuel irradiation capsules, fuel test rods, and steam-generator subsystems~\cite{karnik2024constrained,karnik2024leveraging}. The second mechanism is involuntary sparsity, in which sensors fail, drift outside calibration, or become intermittently unavailable~\cite{woods2019osu,gutowska2023developing}. Both mechanisms create the same reliability challenge: fewer trustworthy observations are available than are needed to directly characterize the complete state. Reconstruction can therefore serve as a state-estimation layer that recovers the full field from the available subset. Here, full-field reconstruction refers specifically to recovering the complete spatially distributed thermocouple response across the instrumented HTTF domain from sparse measurements.

Classical spatial-interpolation and regression methods provide a direct approach to this problem. Radial basis function interpolation, kriging, and deterministic regression estimate values at unobserved locations from available measurements~\cite{laslett1994spatial,stein1999interpolation}. These methods are computationally efficient, interpretable, and well established for scattered spatial data, but their performance depends strongly on sensor coverage and assumptions embedded in the selected kernel, covariance model, or regression form. Standard formulations also do not explicitly represent nonlinear, regime-dependent, or long-range spatiotemporal relationships~\cite{fukami2021global,zhao2024recfno}. This limitation is consequential in transient safety-critical systems, where relationships between unobserved states and available measurements can evolve substantially during an event.

Deterministic machine-learning methods extend classical interpolation by learning nonlinear mappings directly from sparse observations to the missing field. General-purpose reconstruction frameworks have demonstrated strong performance in fluid mechanics and other physical domains. Voronoi tessellation has been used to transform irregular sensor measurements into structured representations for convolutional neural networks~\cite{fukami2021global}, while recurrent networks~\cite{burnett2026variational,farha2025sensor}, graph neural networks (GNNs), and recurrent decoders exploit temporal histories and relational structure in distributed sensor networks~\cite{cini2021filling,marisca2022learning,uremovic2025contextualized}. Within nuclear applications, Gong et al.~\cite{gong2024reactor} reconstructed reactor flux fields from sparse and movable in-core measurements, Batikh et al.~\cite{batikh2025autoencoders} used latent representations to recover temperature gradients during accident and critical power-excursion conditions, and reduced-order sensing has been investigated for temperature-field reconstruction from strategically selected sensors~\cite{price2024simplified}. Complementary studies have forecast accident transients using recurrent networks~\cite{radaideh2020neural} and applied active learning to abnormal-condition analysis and physics-informed surrogate development~\cite{price2023active,nabila2026active,nabila2026physics}. These studies establish the value of nonlinear spatiotemporal representations, but reconstruction robustness when the physical condition differs from training remains less established. This issue is increasingly recognized in reliability research, where domain-generalization methods have been investigated to maintain diagnostic performance under previously unseen working conditions~\cite{azari2025self}.

Generative modeling provides a natural extension by learning a conditional distribution of the missing field given available observations rather than only a deterministic mapping from `known'' to `unknown'' sensors. Variational autoencoders (VAEs) learn latent representations from which missing quantities can be reconstructed probabilistically~\cite{kingma2013auto,diederik2019introduction}, with extensions addressing incomplete tabular and time-series data~\cite{nazabal2020handling,mattei2019miwae,fortuin2020gp,tian2023dual}. Conditional generative adversarial networks (GANs) instead use adversarial training and have been applied to missing-data imputation and energy-system augmentation~\cite{nabila2025data}, power-grid condition prediction~\cite{nabila2025predicting}, and spatiotemporal missing-data recovery~\cite{zhang2023spatiotemporal,gao2022generative,wu2021g}. Together, these approaches motivate probabilistic reconstruction when the available observations do not uniquely determine the unobserved state.

Diffusion models (DMs) provide another probabilistic formulation by learning to reverse a progressive noise-corruption process~\cite{sohl2015deep}. Recent studies have applied DMs to conditional reconstruction and imputation in energy, time-series, and tabular settings~\cite{zhang2025diffusion,tashiro2021csdi,villaizan2025diffusion}. Graph structure, attention, and spatiotemporal priors have improved reconstruction under high missing rates and complete sensor outages~\cite{chen2026temporally}, while physics-based consistency has been introduced for scientific data assimilation under sparse and noisy observations~\cite{wang2025phyda}. These properties make DMs relevant to state reconstruction under degraded sensing, yet their application to safety-critical transient thermal-hydraulic measurements, particularly under complete sensor holdout and transfer between physically distinct experiments, remains limited.

Taken together, the literature exposes a broader reliability question beyond any individual reconstruction architecture: \textit{to what extent can data-driven state reconstruction preserve dependable situational awareness when a safety-critical system becomes partially observable?} Existing surveys of artificial intelligence in nuclear technology~\cite{arhouni2025artificial} demonstrate the expanding role of machine learning but provide limited treatment of conditional state reconstruction under degraded sensing. Three gaps are particularly relevant. First, most nuclear field-reconstruction studies rely on deterministic models and point predictions, leaving generative approaches under partial observability less characterized. Second, reconstruction is commonly evaluated within a single dataset or operating regime, leaving robustness across physically distinct transient conditions insufficiently understood. Third, many studies consider randomly missing entries or a fixed sensor configuration rather than complete sensor-level holdout and progressively reduced sensor availability. Addressing these gaps requires evaluating reconstruction not only as a prediction-accuracy problem but also as a means of preserving state information as sensing conditions become increasingly adverse. The HTTF thermocouple network and its physically distinct transient experiments provide a representative test environment for this investigation through controlled comparisons of deterministic and conditional generative reconstruction methods.

Motivated by these gaps, the present work develops and rigorously benchmarks a suite of conditional generative reconstruction methods---a graph-temporal DM, a conditional VAE, and a conditional GAN---against two deterministic neural-network baselines and a ridge-regression baseline for sparse thermocouple reconstruction across HTTF experiments. All three generative methods are conditioned identically on known thermocouple measurements, three-dimensional sensor coordinates, and transient temperature histories, ensuring that observed performance differences reflect the reconstruction paradigm rather than inconsistencies in conditioning information. In the intra-experiment analysis, multiple known-target sensor permutations are evaluated separately in order for each generative model to find the location of the most informative known sensors. Unlike prior nuclear sparse-reconstruction studies, which report results obtained from a single fixed architecture and sensor layout, this work evaluates six methods under two complementary, increasingly demanding protocols, each with a strict sensor-level holdout: an intra-experiment protocol, which evaluates sensor reconstruction within PG-29 (depressurized conduction cooldown [DCC]), and an inter-experiment protocol, which evaluates cross-experiment reconstruction from PG-27 (pressurized conduction cooldown [PCC]) to PG-29 (DCC), using a section-context architecture and a composite transfer loss. The known-sensor sets selected by the three generative models are also compared so as to identify which sensors contribute most to each model and to examine how their spatial distributions differ. Finally, the sensitivity of model performance to progressively reduced known-sensor counts (91 down to 45) is also studied. The main contributions of this work are as follows:

\begin{enumerate}
    \item \textbf{Systematic reconstruction under progressive sensor loss.} An intra-experiment benchmark within PG-29 compares six reconstruction methods: three GNN--gated recurrent unit (GRU) generative models (DM, VAE, and GAN), two deterministic neural baselines (GNN--GRU and GRU), and ridge regression. The benchmark further evaluates known-sensor locations and quantifies reconstruction robustness as the number of available sensors is progressively reduced from 91 to 45.

    \item \textbf{Reconstruction across physically distinct transient conditions.} An inter-experiment transfer benchmark evaluates whether learned reconstruction relationships remain effective under a change in system conditions by training on PG-27 PCC and reconstructing PG-29 DCC. Section-context DM, VAE, GAN, and GRU models are compared with ridge regression under this cross-transient setting.

    \item \textbf{Sensor-level characterization of reconstruction reliability.} A sensor-wise diagnostic analysis quantifies how reconstruction error varies across thermocouple locations and identifies where generative models provide the largest improvements over regression, providing spatial insight into reconstruction performance beyond aggregate error metrics.
\end{enumerate}

The remainder of this paper is organized as follows. Section~\ref{sec:data} describes the HTTF experimental setup and two experimental conditions. Section~\ref{sec:method} details the theoretical foundations and methodologies employed for model formulation and evaluation. Section~\ref{sec:results} presents a comparative analysis of the generative and non-generative methods. Finally, Section~\ref{sec:conc} summarizes the key contributions and future research directions.

\begin{table*}[!t]  
\small
\begin{framed}
\begin{footnotesize} 

\nomenclature{DM}{Diffusion model}
\nomenclature{VAE}{Variational autoencoder}
\nomenclature{GAN}{Generative adversarial network}
\nomenclature{GNN}{Graph neural network}
\nomenclature{GRU}{Gated recurrent unit}
\nomenclature{PCC}{Pressurized conduction cooldown}
\nomenclature{DCC}{Depressurized conduction cooldown}
\nomenclature{RCCS}{Reactor cavity cooling system}
\nomenclature{RCST}{Reactor cavity simulation tank}
\nomenclature{HTTF}{High Temperature Test Facility}
\nomenclature{TS}{Temperature solid}
\nomenclature{MAE}{Mean absolute error}
\nomenclature{RMSE}{Root mean square error}

\end{footnotesize}

\printnomenclature

\end{framed}

\end{table*}

\section{Experimental Facility and Data} 
\label{sec:data}

\subsection{High Temperature Test Facility at Oregon State University} \label{sec:httf}

The Oregon State University HTTF is a reduced-scale integral experimental facility designed to reproduce the dominant thermal-hydraulic behavior of modular high-temperature gas-cooled reactors~\cite{woods2019instrumentation,woods2019osu,gutowska2023developing}. The system implements a geometric scale of one quarter in height and diameter and an approximate pressure scale of one eighth relative to the reference reactor. These similarity relationships allow the HTTF to achieve prototypic temperature distributions while maintaining material compatibility and experimental accessibility. The ceramic blocks comprising the surrogate core preserve the thermal characteristics of graphite at scaled dimensions, and the internal structure includes more than 500 coolant channels and over 200 heater channels arranged to emulate the flow and heat-transfer pathways of the modular high-temperature gas-cooled reactor prismatic core.

Instead of nuclear fuel, the facility employs electrically heated graphite rodlets inserted into designated heater channels. These rodlets serve as distributed heat sources capable of reproducing decay-heat profiles under controlled experimental conditions. The heaters are organized into multiple banks located below and within the core, enabling fine control of the axial and radial power distribution. With full heater capability, the system can deliver up to approximately 2.2~MW of thermal power; however, for depressurized cooldown experiments, the facility typically operates at reduced power levels below 350~kW so as to represent scaled decay heat while also preventing thermal overstress of the ceramic structures.

The integral nature of the HTTF permits a realistic representation of reactor vessel and primary-loop behavior. Helium circulates through the lower plenum, upward through the core channels, and into the upper plenum during forced convection, before returning through the concentric crossover-ducts. The vessel is enclosed within a reactor cavity that houses a scaled reactor cavity cooling system (RCCS), which provides passive heat rejection through radiation and natural convection to the cavity environment. This configuration enables the facility to capture the complex transition from forced flow to conduction- and buoyancy-dominated heat removal mechanisms.

Extensive instrumentation is embedded throughout the core, plena, crossover-ducts, and cavity structures to measure solid and fluid temperatures during transient operation. These measurements allow detailed reconstruction of flow reversal, stratification, lock-exchange behavior between helium and nitrogen, and the onset of natural circulation following depressurization. As a result, the HTTF represents a high-fidelity experimental platform for validating thermal-hydraulic models, assessing passive safety performance, and generating benchmark datasets for advanced diagnostics and DT research.

\subsection{Depressurized Conduction Cooldown (PG\textendash29) Test Description}
\label{sec:dcc}

The PG{-}29 experiment was a low-power integral DCC test conducted at the Oregon State University HTTF to investigate depressurization, gas exchange, thermal stratification, and natural circulation following a postulated double-ended inlet--outlet crossover-duct break~\cite{gutowska2023developing}. The test was performed from July~24 to 26, 2019, using the hybrid heater configuration at power levels below 350~kW. Following facility heatup, the average core temperature was approximately 550$^\circ$C--590$^\circ$C, with peak temperatures of approximately 780$^\circ$C--820$^\circ$C. The primary system initially contained pressurized helium, whereas the adjacent reactor cavity simulation tank (RCST) contained nitrogen at a slightly lower pressure. Thus, ``depressurized'' refers to the transient following break opening rather than to the initial primary-system condition. The water-filled RCCS remained a separate heat-removal system.

The DCC transient began slightly after 80~h with shutdown of the primary circulator, immediately eliminating forced convection, followed approximately a minute later by opening the concentric hot- and cold-leg break valves. The resulting depressurization produced counter-current gas exchange, with hot helium flowing from the reactor pressure vessel into the RCST and cooler nitrogen entering the primary system, thereby establishing an initially stratified flow regime. Subsequent heat redistribution was governed primarily by conduction within the ceramic structures and buoyancy-driven natural circulation. An unplanned trip of the hybrid heater bank at near 89~h terminated active heating earlier than intended, after which cooling proceeded via solid conduction, gas-phase natural convection, and radiative heat transfer to the RCCS panels. The reported transient concluded when RCCS operation was terminated at approximately 95.5~h.

\subsection{Pressurized Conduction Cooldown (PG\textendash27) Test Description}
\label{sec:pcc_pg27}

The PG{-}27 experiment was a low-power integral PCC test conducted at the HTTF to investigate heat removal following loss of forced circulation while the primary system remained pressurized. The overall campaign extended from April~23 to May~24,~2019; however, only the final PG27{-}G dataset is used in this study, as earlier test segments were interrupted by water-related reductions in heater-cable insulation resistance. The experiment used heater banks 102 and 108 in tandem with a scaled decay-power profile below 350~kW. Before PCC initiation, the primary system and RCST contained helium at pressures above approximately 130 and 101~kPa, respectively, while the water-filled RCCS remained available for cavity heat removal.

During the final PG{-}27 test sequence, heater power was increased gradually toward the prescribed decay-power condition but was limited by ceramic temperature and heatup-rate constraints. Near PCC initiation, each heater bank supplied approximately 52~kW, with later adjustments being made so as to remain within facility limits. The PCC transient began approximately 69~h after the start of the test, when forced circulation was stopped and the steam generator was isolated. The heated PCC phase ended approximately 73~h and 29~min after test initiation, when water was detected in the low-point drains and deterioration of the heater insulation required the heaters to be shut down, after which the facility entered an unheated cooldown period. Although this premature termination altered the nominal long-duration PCC trajectory, the test still captured the early heatup, PCC initiation, decay-power operation, and subsequent unheated cooldown behavior relevant to low-power HTTF transient analysis. The main differences between PG{-}27 and PG{-}29 are summarized in Table~\ref{tab:pg27_pg29_key_features}.

\begin{table}[H]
\centering
\caption{Key operating characteristics of the PG{-}29 (DCC)
and PG{-}27 (PCC) experiments used in this study.}
\label{tab:pg27_pg29_key_features}
\renewcommand{\arraystretch}{1.10}
\setlength{\tabcolsep}{4pt}
\footnotesize
\begin{tabular}{
    p{0.23\columnwidth}
    p{0.355\columnwidth}
    p{0.355\columnwidth}}
\toprule
\textbf{Feature}
& \textbf{PG{-}29 DCC}
& \textbf{PG{-}27 PCC} \\
\midrule

Transient type
& DCC
& PCC \\

Postulated scenario
& Double-ended inlet--outlet crossover-duct break
& Loss of forced circulation under pressurized conditions \\

Reported test period
& July 24--26, 2019
& April 23--May 24, 2019 \\

Data window used
& 45.6~h
& 93.4~h \\

Role in the present study
& Intra-experiment reconstruction and
inter-experiment evaluation
& Training of the inter-experiment
reconstruction models \\

\bottomrule
\end{tabular}
\end{table}

\subsection{Data Preparation}
\label{sec:predata}

\begin{figure}[!h]
    \centering
    \includegraphics[width=0.7\linewidth]{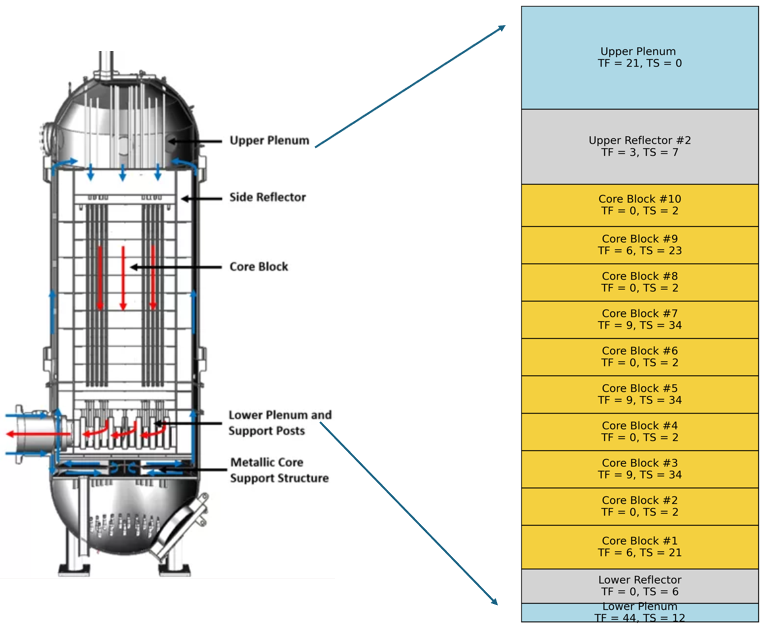}
    \caption{Axial layout of the HTTF vessel, and distribution of TF and TS thermocouples across the major facility regions.}
    \label{fig:httf_axial}
\end{figure}

Fig.~\ref{fig:httf_axial} illustrates axial segmentation of the HTTF vessel and the distribution of fluid-temperature (TF) and solid-temperature (TS) thermocouples across the upper plenum, upper reflector, 10 core blocks, lower reflector, and lower plenum.

The HTTF test reports identify several non-functioning or intermittently unavailable thermocouple channels in the two experiments. The original HTTF instrumentation layout included 155~TF and 155~TS thermocouples, for a total of 310 sensors, providing extensive spatial coverage of coolant and solid temperatures throughout the facility. Only the TS measurements are used in the present study. When a valid backup measurement was available, missing portions of the corresponding signal were recovered through numerical interpolation. Channels that could not be reconstructed reliably from either the primary or backup instrumentation were excluded. This preprocessing resulted in 149 usable TS thermocouples for the intra-experiment analysis, of which 137 were available in both PG{-}27 and PG{-}29 and were retained for the inter-experiment analysis. A 93.4~h window from PG{-}27 and a 45.6~h window from PG{-}29 were used for sparse-sensor reconstruction. Both datasets were resampled to a uniform 30~s interval.

\section{Methodology} 
\label{sec:method}

\begin{figure}[!h]
    \centering
    \includegraphics[width=0.9\textwidth]{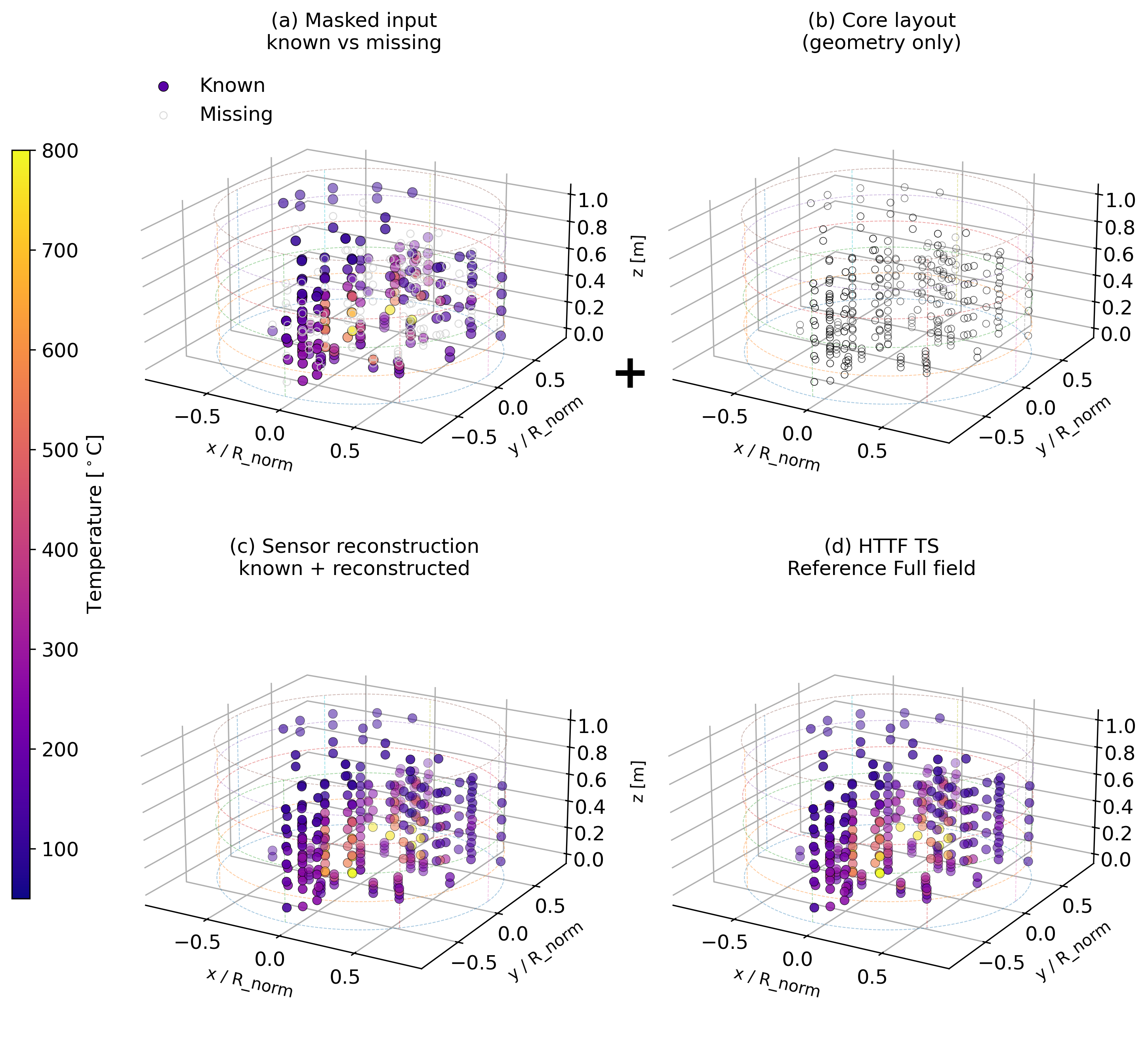}
    \caption{Sparse-sensor reconstruction workflow for the HTTF thermocouple field.}
    \label{fig:DM_framework}
\end{figure}

Fig.~\ref{fig:DM_framework} summarizes the generative-model-based reconstruction procedure used in this study. Panel~(d) illustrates the full HTTF temperature field at a representative time step, with the sensor locations shown in three-dimensional space and colored in accordance with the measured temperature.

To emulate sparse instrumentation, a subset of sensors is designated as ``known'' while the remaining sensors are masked and treated as ``unknown,'' as shown in Panel~(a). The known sensors provide the conditioning measurements for the DM, while the masked locations represent temperatures to be reconstructed. Importantly, the spatial distribution of known and unknown sensors reflects realistic, nonuniform sensor coverage rather than idealized
regular grids.

Panel~(b) shows the geometric configuration of the sensor network alone, without any temperature information. These normalized spatial coordinates are explicitly provided to the DM as part of the conditioning vector, enabling the model to learn spatial correlations and geometry-aware temperature patterns across the reactor vessel.

Panel~(c) shows the final reconstructed field obtained by combining the measured temperatures at the known sensors with the model-reconstructed values at the unknown locations. The reconstructed field preserves global axial and radial temperature trends while smoothly interpolating across masked regions. This end-to-end workflow demonstrates how conditional generative models enable full-field temperature reconstruction from sparse, irregular sensor measurements, using only sensor values and geometric information.

This framework is shared by the DM, conditional VAE, and conditional GAN considered in this work. The models differ in their generative formulations and training objectives, but all rely on the same fundamental reconstruction task: estimating a spatially complete and temporally consistent HTTF temperature field from sparse, irregularly distributed sensor observations. The following evaluation protocols define how the known, training-target, and fixed test sensors are separated under the intra- and inter-experiment settings.

\subsection{Evaluation Protocol}
\label{subsec:evaluation_protocol}

Two complementary evaluation protocols are employed to evaluate sparse thermocouple reconstruction under progressively more challenging generalization settings. The first protocol assesses reconstruction within a single HTTF experiment, whereas the second evaluates the ability of models trained on one transient to generalize to a different experiment. Across both protocols, all methods are evaluated using identical known and hidden sensor sets so as to ensure a fair comparison. 

Six reconstruction methods are investigated. Three are generative AI models based on latent-variable or diffusion formulations, while the rest are deterministic neural networks and a regression baseline. The three generative models are specifically designed to capture increasingly rich spatial and temporal thermal-hydraulic relationships, whereas the regression model provides a conventional non-generative reference. Each model is introduced in detail in the following subsections.

\begin{table}[!t]
\centering
\caption{Data assignments for the intra- and inter-experiment reconstruction protocols.}
\label{tab:data_split}
\small
\begin{tabular}{lccccc}
\hline
\textbf{Protocol} &
\textbf{Total sensors} &
\textbf{Training data} &
\textbf{Known sensors} &
\textbf{Training targets} &
\textbf{Test targets} \\
\hline
Intra-experiment &
149 &
PG{-}29 &
91 &
29 PG{-}29 &
29 PG{-}29 \\
Inter-experiment &
137 &
PG{-}27 &
111   &
26 PG{-}27 &
26 PG{-}29 \\
\hline
\end{tabular}
\end{table}

\textbf{Intra-experiment reconstruction:}
The intra-experiment protocol evaluates sparse-sensor reconstruction within PG{-}29 (DCC) while enforcing a strict sensor-level holdout. From the 149 available TS thermocouples, 29 sensors (20\% of the total) are assigned to a fixed test set. The test sensors were selected to be mutually distant and spatially distributed across the instrumented HTTF domain, providing a representative sparse-sensor reconstruction task while simultaneously avoiding bias stemming from random sensors being concentrated in a single corner or local cluster. These fixed test sensors are excluded from model inputs and training loss and are used only for final evaluation. The remaining sensors are divided into 91 known conditioning sensors---corresponding to 60\% of the total sensor set---and 29 sensors used for training purposes. For this setting, the models are implemented as a GNN--GRU-generative architecture. Graph layers aggregate information among neighboring thermocouples, while a GRU processes the graph embeddings over the temporal window. This architecture allows the model to use both the physical sensor layout and transient temperature history when reconstructing the 29 held-out sensors. 

All three generative methods use the same types of conditioning information, including known thermocouple measurements, three-dimensional sensor coordinates, and transient temperature histories. Each model was evaluated over multiple random known--target sensor permutations, and the selected cases therefore use partially different known-sensor sets. Their overlap and spatial distributions are examined to assess the possible influence of conditioning coverage on the reported comparison. Table~\ref{tab:data_split} summarizes the sensor assignments and training/test data used for the intra-experiment reconstruction protocol.

\textbf{Inter-experiment reconstruction:}
The inter-experiment protocol evaluates transferability across HTTF experiments by training on PCC data and testing on DCC data. Among the fixed missing sensors, 26 thermocouples are available in both experiments and are used as the fixed inter-experiment test targets. The remaining common thermocouples are treated as known sensors for conditioning and baseline reconstruction. For this setting, the generative models are implemented as a section-context generative architecture. Each section is defined as a group of thermocouples with similar PCC transient behavior, including peak timing, heating and cooling rates, and segmented transient trends. The inter-experiment loss extends the main diffusion loss with temperature-space, local-neighbor, section-context, transient-slope, and residual-magnitude penalties to improve PCC-to-DCC transfer. Thus, this protocol tests whether reconstruction knowledge learned from one HTTF transient can generalize to another without using DCC missing-sensor truth during training. Table~\ref{tab:data_split} summarizes the sensor assignments and training/test data used for the inter-experiment reconstruction protocol.

\subsection{Conditional Diffusion Model}
\label{subsec:dm}

DMs are probabilistic generative models that learn complex data distributions by reversing a gradual noise-corruption process~\cite{sohl2015deep,ho2020denoising}. During training, clean data are progressively perturbed with Gaussian noise, while a neural denoiser learns to recover the original signal from increasingly corrupted observations. For sparse thermocouple reconstruction, this process is formulated as a conditional generation problem in which the model reconstructs the missing thermocouple temperatures while conditioning on the available sensor measurements, sensor locations, and transient history~\cite{tashiro2021csdi}.

Let $\mathbf{z}_0$ denote the normalized target temperature field corresponding to the hidden thermocouples. The forward diffusion process gradually corrupts the target by adding Gaussian noise:

\begin{equation}
\mathbf{z}_s
=
\sqrt{\bar{\alpha}_s}\mathbf{z}_0
+
\sqrt{1-\bar{\alpha}_s}\boldsymbol{\epsilon},
\qquad
\boldsymbol{\epsilon}
\sim
\mathcal{N}(\mathbf{0},\mathbf{I}),
\end{equation}
where $s$ denotes the diffusion timestep and $\bar{\alpha}_s$ is the cumulative noise schedule. As the diffusion process progresses, the target becomes increasingly noisy, forcing the model to learn the underlying temperature distribution rather than memorizing individual sensor values.

The reverse process is parameterized by a conditional neural denoiser:

\begin{equation}
\widehat{\boldsymbol{\epsilon}}
=
\epsilon_{\theta}
(
\mathbf{z}_s,
\mathbf{h}_t,
s
),
\end{equation}
where $\mathbf{h}_t$ denotes the conditioning information obtained from the available thermocouple measurements, sensor-location information, and temporal context. The predicted noise is then used to estimate the clean reconstruction:

\begin{equation}
\widehat{\mathbf{z}}_0
=
\frac{
\mathbf{z}_s
-
\sqrt{1-\bar{\alpha}_s}
\widehat{\boldsymbol{\epsilon}}
}{
\sqrt{\bar{\alpha}_s}
}.
\end{equation}

The DM is trained using a combination of the standard noise-prediction objective and a direct reconstruction objective:

\begin{equation}
\mathcal{L}_{\mathrm{DM}}
=
\lambda_{\epsilon}
\left\|
\boldsymbol{\epsilon}_{\mathcal{S}_{\mathrm{train}}}
-
\widehat{\boldsymbol{\epsilon}}_{\mathcal{S}_{\mathrm{train}}}
\right\|_2^2
+
\lambda_{0}
\left\|
\mathbf{z}_{0,\mathcal{S}_{\mathrm{train}}}
-
\widehat{\mathbf{z}}_{0,\mathcal{S}_{\mathrm{train}}}
\right\|_2^2,
\end{equation}
where the first term teaches the network to reverse the diffusion process and the second term directly improves the reconstruction accuracy of the hidden thermocouples. The optimization is performed only over the train-target sensors, while the fixed test sensors remain completely excluded from both the conditioning inputs and the training loss.

During inference, reconstruction begins from Gaussian noise and iteratively applies the learned reverse diffusion process:

\begin{equation}
\mathbf{z}_{s-1}
=
\boldsymbol{\mu}_{\theta}
(
\mathbf{z}_s,
\mathbf{h}_t,
s
)
+
\sigma_s\boldsymbol{\eta},
\qquad
\boldsymbol{\eta}
\sim
\mathcal{N}(\mathbf{0},\mathbf{I}),
\end{equation}
where $\boldsymbol{\mu}_{\theta}$ is the reverse-process posterior mean computed from the predicted noise, $\sigma_s$ is the timestep-dependent noise variance, and the stochastic term is omitted at the final denoising step. Multiple generated samples may be averaged to reduce sampling variability and improve reconstruction stability.

For the intra-experiment protocol, the DM employs a GNN to aggregate spatial information among neighboring thermocouples and a GRU to encode transient temperature evolution before conditional diffusion-based reconstruction. For the inter-experiment protocol, the same conditional diffusion framework is extended with section-context conditioning and a composite loss that incorporates local-neighbor consistency, temperature-space similarity, transient-slope matching, and residual-magnitude regularization. These additions encourage the model to transfer thermal-hydraulic relationships learned from the PG{-}27 PCC experiment to the PG{-}29 DCC experiment while reconstructing thermocouples that remain unseen during training.

\subsection{Conditional Variational Autoencoder}
\label{subsec:vae}

VAEs are probabilistic generative models that learn compact latent representations by combining neural encoding, stochastic latent sampling, and neural decoding
\cite{kingma2013auto,diederik2019introduction}. Unlike DMs, which generate a reconstruction through an iterative denoising process, a VAE maps the available sensor observations into a lower-dimensional latent space and decodes the sampled latent representation directly into the missing thermocouple responses.

Let $\mathbf{x}_t \in \mathbb{R}^{N}$ denote the normalized thermocouple field at time $t$, where $N$ is the total number of sensors. Let $\mathbf{m}\in\{0,1\}^{N}$ denote the observation mask, with $m_i=1$ for a known sensor and $m_i=0$ for a hidden sensor. The masked observation is:

\begin{equation}
\widetilde{\mathbf{x}}_t
=
\mathbf{m}\odot\mathbf{x}_t,
\end{equation}
where $\odot$ denotes element-wise multiplication. For a temporal window of length $L$, the conditioning representation is constructed as:

\begin{equation}
\mathbf{h}_t
=
g_{\psi}
\left(
\widetilde{\mathbf{X}}_{t-L+1:t},
\mathbf{m},
\mathbf{c}
\right),
\end{equation}
where $\widetilde{\mathbf{X}}_{t-L+1:t}$ contains the masked thermocouple measurements over the temporal window, $\mathbf{c}$ contains the normalized three-dimensional sensor coordinates, and $g_{\psi}$ denotes the graph-temporal conditioning network. Thus, $\mathbf{h}_t$ encodes the known temperature history, observation mask, sensor geometry, and transient context without using the hidden-sensor values as conditioning inputs.

The variational encoder maps the conditioning representation to a diagonal Gaussian latent posterior:

\begin{equation}
q_{\phi}
\left(
\mathbf{z}\mid\mathbf{h}_t
\right)
=
\mathcal{N}
\left(
\boldsymbol{\mu}_{\phi}(\mathbf{h}_t),
\operatorname{diag}
\left[
\boldsymbol{\sigma}_{\phi}^{2}(\mathbf{h}_t)
\right]
\right),
\end{equation}
where $\boldsymbol{\mu}_{\phi}$ and $\boldsymbol{\sigma}_{\phi}$ are the latent mean and standard-deviation vectors predicted by the encoder. In practice, the encoder may predict the logarithm of the latent variance, $\log\boldsymbol{\sigma}_{\phi}^{2}$, from which the
standard deviation is obtained as:

\begin{equation}
\boldsymbol{\sigma}_{\phi}
=
\exp
\left(
\frac{1}{2}
\log\boldsymbol{\sigma}_{\phi}^{2}
\right).
\end{equation}

A latent sample is drawn using the reparameterization trick:

\begin{equation}
\mathbf{z}
=
\boldsymbol{\mu}_{\phi}
+
\boldsymbol{\sigma}_{\phi}
\odot
\boldsymbol{\epsilon},
\qquad
\boldsymbol{\epsilon}
\sim
\mathcal{N}(\mathbf{0},\mathbf{I}),
\end{equation}
which separates the stochastic sampling operation from the trainable encoder parameters and permits gradient-based optimization.

The conditional decoder reconstructs the thermocouple field from the latent sample and the conditioning representation:

\begin{equation}
\widehat{\mathbf{x}}_t
=
f_{\theta}
\left(
\mathbf{z},
\mathbf{h}_t
\right),
\end{equation}
where $f_{\theta}$ denotes the decoder network. The decoder predicts a complete candidate field, but the reconstruction loss is evaluated only at the designated train-target sensors. The known measurements are preserved when the final completed field is assembled:

\begin{equation}
\mathbf{x}^{\mathrm{comp}}_t
=
\mathbf{m}\odot\mathbf{x}_t
+
(\mathbf{1}-\mathbf{m})
\odot
\widehat{\mathbf{x}}_t.
\end{equation}

Therefore, measured values are retained exactly at the known sensor locations, while model-generated values are inserted only at hidden locations.

The VAE is trained using a weighted combination of reconstruction and Kullback--Leibler (KL) divergence losses:

\begin{equation}
\mathcal{L}_{\mathrm{VAE}}
=
\lambda_{\mathrm{rec}}
\frac{1}{N_{\mathrm{train}}}
\left\|
\mathbf{x}_{t,\mathcal{S}_{\mathrm{train}}}
-
\widehat{\mathbf{x}}_{t,\mathcal{S}_{\mathrm{train}}}
\right\|_2^2
+
\lambda_{\mathrm{KL}}
D_{\mathrm{KL}}
\left[
q_{\phi}
\left(
\mathbf{z}\mid\mathbf{h}_t
\right)
\,\middle\|\,
\mathcal{N}(\mathbf{0},\mathbf{I})
\right],
\end{equation}
where $\mathcal{S}_{\mathrm{train}}$ denotes the train-target sensor set and $N_{\mathrm{train}}=|\mathcal{S}_{\mathrm{train}}|$. The first term is the mean squared reconstruction error over the train-target sensors. The second term regularizes the approximate posterior toward a standard normal prior, promoting a smooth and structured latent space.

For the diagonal Gaussian posterior, the KL term has the closed-form expression:

\begin{equation}
D_{\mathrm{KL}}
\left[
q_{\phi}
\left(
\mathbf{z}\mid\mathbf{h}_t
\right)
\,\middle\|\,
\mathcal{N}(\mathbf{0},\mathbf{I})
\right]
=
-\frac{1}{2}
\sum_{j=1}^{d_z}
\left[
1
+
\log \sigma_{\phi,j}^{2}
-
\mu_{\phi,j}^{2}
-
\sigma_{\phi,j}^{2}
\right],
\end{equation}
where $d_z$ is the latent dimension. The weighting coefficients
$\lambda_{\mathrm{rec}}$ and $\lambda_{\mathrm{KL}}$ control the balance
between reconstruction accuracy and latent-space regularization.

During training, the values at the train-target sensors are used only to calculate the reconstruction loss, and are not included in the conditioning measurements. The fixed test sensors are excluded from both the conditioning inputs and the training loss and are retained only for final evaluation. During inference, the encoder receives only the available thermocouple measurements, observation mask, sensor coordinates, and temporal history. The
decoder then generates the missing sensor values from the resulting latent representation.

For the intra-experiment protocol, the VAE employs graph-based spatial encoding and temporal attention to learn spatial and transient correlations directly from PG{-}29. For the inter-experiment protocol, the same conditional VAE framework is extended with section-context information derived from PG{-}27 transient characteristics, enabling latent representations that better transfer across different HTTF experiments.

\subsection{Conditional Generative Adversarial Network}
\label{subsec:gan}

GANs learn a data distribution through competition between two neural networks: a generator and a discriminator~\cite{goodfellow2020generative,nabila2025data}. The generator tries to produce reconstructed temperature fields that look like measured HTTF data, while the discriminator tries to determine whether each field is real or generated. As training progresses, the generator improves by learning to fool the discriminator. When the discriminator can no longer reliably distinguish generated fields from measured ones, the generator is considered to have learned the main characteristics of the data distribution.

Let $\mathbf{h}_t$ denote the conditioning information constructed from the known thermocouple measurements, observation mask, sensor coordinates, and temporal history. The generator produces the reconstructed field as:

\begin{equation}
\widehat{\mathbf{x}}_t
=
G(\mathbf{h}_t).
\end{equation}

The discriminator receives either a measured or reconstructed field, together with the same conditioning information:

\begin{equation}
D(\mathbf{x}_t,\mathbf{h}_t)\in[0,1].
\end{equation}
It then estimates the probability that the supplied field belongs to the measured HTTF data distribution. The conditional adversarial learning problem is formulated as:

\begin{equation}
\min_{G}\max_{D}
\mathcal{V}(D,G)
=
\mathbb{E}_{(\mathbf{x}_t,\mathbf{h}_t)\sim p_{\mathrm{data}}}
\left[
\log D(\mathbf{x}_t,\mathbf{h}_t)
\right]
+
\mathbb{E}_{\mathbf{h}_t\sim p_{\mathrm{data}}}
\left[
\log
\left(
1-
D\left(G(\mathbf{h}_t),\mathbf{h}_t\right)
\right)
\right].
\end{equation}

To ensure accurate prediction of individual thermocouple responses, the generator is also trained using a supervised reconstruction loss evaluated only over the train-target sensor set. The adversarial component encourages the generated fields to follow the overall distribution of measured HTTF temperatures, while the reconstruction component preserves sensor-specific temperature amplitudes and transient trends. The fixed test sensors are excluded from both the conditioning inputs and training losses and are retained only for final evaluation.

For the intra-experiment protocol, the generator and discriminator use graph-based spatial representations and temporal features to reconstruct hidden PG{-}29 thermocouples. For the inter-experiment protocol, section-context information derived from PG{-}27 is incorporated into the conditioning representation to improve transfer from the PCC experiment to the PG{-}29 DCC
experiment.

% ARCHITECTURE TABLE
\begin{table*}[!t]
\centering
\caption{Architectures and principal hyperparameters of the conditional
generative models.}
\label{tab:model_architecture}
\footnotesize
\setlength{\tabcolsep}{5pt}
\renewcommand{\arraystretch}{1.14}

\begin{tabularx}{\textwidth}{
    >{\raggedright\arraybackslash}p{3.1cm}
    >{\raggedright\arraybackslash}X}
\toprule
\textbf{Model} &
\textbf{Model-specific architecture and hyperparameters} \\
\midrule

% ============================================================
% INTRA-EXPERIMENT
% ============================================================
\multicolumn{2}{l}{
\textbf{Intra-experiment reconstruction: PG{-}29}
} \\
\addlinespace[2pt]

\multicolumn{2}{p{\dimexpr\textwidth-2\tabcolsep\relax}}{
\textit{Shared settings:}
temporal window $L=40$; graph neighbors $k=8$;
conditioning GNN $2\rightarrow64\rightarrow64$;
batch size $64$; one-layer GRU $64\rightarrow256$; gradient clipping $1.0$; Adam optimizer; 
normalization: StandardScaler; $150$ epochs.
} \\
\addlinespace[3pt]

\textbf{GNN--GRU DM}
&
Denoiser input dimension:
$1+1+192+192+256+256=898$.
Denoiser MLP:
$898\rightarrow192\rightarrow192\rightarrow192\rightarrow1$;
SiLU activation; clean-state clipping to $[-5,5]$.
Learning rate $10^{-3}$; $150$ epochs;
$250$ diffusion steps;
$\beta_{\min}=5\times10^{-5}$ and
$\beta_{\max}=10^{-2}$;
$\lambda_c=1.0$ and $\lambda_0=0.5$;
$10$ inference samples.
\\
\cmidrule(lr){1-2}

\textbf{GNN--GRU VAE}
&
Latent dimension $32$;
coordinate encoder
$d_{\mathrm{coord}}\rightarrow64\rightarrow64$.

Per-sensor decoder input dimension:
$32+128+64+1+1=226$.
Decoder:
$226\rightarrow128\rightarrow128\rightarrow1$,
with ReLU activation and dropout $0.10$.
Learning rate $1\times10^{-4}$.
\\
\cmidrule(lr){1-2}

\textbf{GNN--GRU GAN}
&
Noise dimension $32$;
coordinate and global-noise hidden dimensions $192$;
dropout $0.05$.

Generator input dimension:
$1+1+192+192+256+256+32=930$.
Generator MLP:
$930\rightarrow192\rightarrow192\rightarrow192\rightarrow1$,
with SiLU activation.

Discriminator input dimension:
$1+1+1+1+192+256+256=708$.
Discriminator MLP:
$708\rightarrow192\rightarrow192\rightarrow192\rightarrow1$,
with LeakyReLU activation and binary cross-entropy loss.

Learning rate $5\times10^{-4}$;
Adam coefficients $(0.5,0.999)$.
\\

\midrule

% ============================================================
% INTER-EXPERIMENT
% ============================================================
\multicolumn{2}{l}{
\textbf{Inter-experiment reconstruction:
PG{-}27$\rightarrow$PG{-}29}
} \\
\addlinespace[2pt]

\multicolumn{2}{p{\dimexpr\textwidth-2\tabcolsep\relax}}{
\textit{Shared settings:}
Temporal window $L=40$; eight physical neighbors;
six transient sections; one GRU layer;
GRU hidden dimension $128$;
context hidden dimension $256$;
target hidden dimension $128$;
dropout $0.05$;
AdamW optimizer with learning rate $10^{-3}$ and
weight decay $10^{-6}$;
batch size $256$;
$250$ epochs.
} \\
\addlinespace[3pt]

\textbf{Section-context GRU--DM}
&
Time-embedding dimension $96$.
Denoiser input dimension:
$1+256+128+96+7=488$.
Denoiser MLP:
$488\rightarrow256\rightarrow256\rightarrow1$.
$80$ diffusion steps;
$\beta_{\min}=5\times10^{-5}$ and
$\beta_{\max}=10^{-2}$;
$10$ inference samples.
\\
\cmidrule(lr){1-2}

\textbf{Section-context GRU--VAE}
&
Latent dimension: $24$; context vector: $391$.
Conditional prior and posterior: two hidden layers of width $256$; 
Decoder: $415\rightarrow256\rightarrow256\rightarrow1$;
KL weight $\beta_{\mathrm{KL}}=0.015$.
\\
\cmidrule(lr){1-2}

\textbf{Section-context GRU--GAN}
&
Noise dimension $32$.

Generator input dimension:
$32+256+128+7=423$. Generator MLP:
$423\rightarrow256\rightarrow256\rightarrow1$,
with SiLU activation. Discriminator input dimension:
$1+256+128+7=392$.
Discriminator MLP:
$392\rightarrow256\rightarrow256\rightarrow1$,
with LeakyReLU activation.
\\

\bottomrule
\end{tabularx}
\end{table*}

Table~\ref{tab:model_architecture} summarizes the architectures and principal hyperparameters used by the three conditional generative models under the intra- and inter-experiment protocols. For intra-experiment reconstruction, the DM, VAE, and GAN use GNN--GRU conditioning backbones that combine graph-based spatial encoding with temporal modeling. The deterministic GNN--GRU baseline uses the same underlying graph and recurrent hyperparameters as the corresponding intra-experiment generative architectures, but removes the diffusion, latent-variable, or adversarial components. The plain GRU baseline retains the same temporal-window and recurrent-model settings while excluding graph-based spatial encoding. For inter-experiment reconstruction, the section-context GRU baseline uses the same physical-context, section-context, temporal-window, and GRU backbone settings as the section-context generative models, while the plain GRU baseline excludes the section-context inputs. These shared base configurations allow the contribution of the generative formulation and spatial-context conditioning to be assessed more directly.

\subsection{Baseline Models}
\label{subsec:baseline}

Three non-generative baselines are considered for sparse thermocouple reconstruction: ridge regression, a GRU-based temporal model, and a deterministic spatiotemporal model. All baselines are evaluated using the same known, train-target, and fixed-test sensor sets as the generative models, ensuring a fair comparison under identical sparse-sensor conditions. For intra-experiment reconstruction, the deterministic GNN--GRU baseline retains the same graph-based spatial encoder and GRU temporal backbone used by the generative models, but removes the latent, adversarial, or diffusion-based formulation. A plain GRU baseline is also included to evaluate temporal reconstruction without graph-based spatial encoding.

For inter-experiment reconstruction, a deterministic section-context GRU is included to isolate the contribution of the generative formulation under PG{-}27-to-PG{-}29 transfer. This model combines section-context conditioning with a GRU backbone, while a plain GRU baseline evaluates temporal reconstruction without section-context information.

The ridge-regression baseline predicts each hidden thermocouple from the available known sensors while using \(L_2\) regularization to reduce the instability caused by strong correlations among thermocouple signals \cite{sasaki2025fast}. The GRU baseline models temporal dependencies in the known-sensor histories without explicit spatial message passing. The GNN--GRU baseline combines graph-based inter-sensor information with recurrent temporal encoding, providing a deterministic counterpart to the conditioning architecture used by the DM.

Additional approaches, including radial-basis-function interpolation, unregularized linear regression, and a GNN-only model, were also examined during preliminary testing. These methods produced worse reconstruction performance and are therefore not included in the final comparison, due to space constraints.

\section{Results}
\label{sec:results}

The results are organized into three main parts. First, the intra-experiment analysis evaluates reconstruction performance within the same transient, including global and sensor-wise accuracy, transient reconstruction behavior, computational cost, known-sensor placement, and sensitivity to reducing the number of available sensors. Second, the inter-experiment analysis evaluates the more challenging transfer from the PG-27 PCC transient to the PG-29 DCC transient to assess generalization across distinct thermal-hydraulic conditions. Finally, the discussion synthesizes the findings from both protocols and examines their broader implications for sparse-sensor reconstruction and deployment.

\subsection{Intra-Experiment Reconstruction}
\label{sec:intra}

Global performance under the intra-experiment protocol is summarized in Table~\ref{tab:intra_metrics}. The GNN--GRU VAE achieves the best overall reconstruction accuracy, with a mean absolute error (MAE) of $11.73\,^\circ$C, an RMSE of $20.54\,^\circ$C, and an $R^2$ of $0.98$. It is followed closely by the GNN--GRU GAN, with an RMSE of $22.37\,^\circ$C, then the GNN--GRU DM, with an RMSE of $22.61\,^\circ$C. The deterministic GNN--GRU and GRU baselines produce higher RMSE values of $25.26\,^\circ$C and $35.22\,^\circ$C, respectively, while ridge regression yields the largest error, with an RMSE of $48.62\,^\circ$C and an $R^2$ of $0.88$. Relative to regression, the VAE, GAN, and DMs reduce the RMSE by approximately 58\%, 54\%, and 53\%, respectively. The deterministic GNN--GRU also outperforms the plain GRU, demonstrating the benefit of incorporating graph-based spatial information in addition to temporal modeling. Overall, the results show that all three generative formulations provide strong and comparable reconstruction performance under the matched intra-experiment setting, with the VAE achieving the lowest aggregate error.

Removing graph-based spatial encoding and retaining the temporal GRU boosts the RMSE from $20.54\,^\circ$C for the best generative model to $35.22\,^\circ$C, an increase of approximately 71\%. This result indicates that temporal information alone is less effective than a combined spatiotemporal representation for reconstructing the distributed HTTF temperature field. The remaining gap between the deterministic GNN-GRU backbone and its three generative counterparts (RMSE $25.26\,^\circ$C vs.\ $20.54$--$22.61\,^\circ$C) is smaller but still consistent and non-negligible, indicating that layering a generative, distributional formulation---whether variational, adversarial, or diffusion-based---on top of the same GNN-GRU backbone yields a further, reproducible improvement beyond what the deterministic spatiotemporal architecture achieves on its own. 

Two points merit emphasis beyond a simple ranking. First, the spread among the three generative models ($\approx2\,^\circ$C RMSE) is small relative to the gap separating any of them from regression ($\approx26\,^\circ$C), suggesting that the dominant source of improvement is the generative, spatially conditioned modeling paradigm itself rather than the specific choice of VAE, GAN, or diffusion formulation. Second, the DM's aggregate metrics are essentially tied with those of the GAN, despite the DM alone offering a principled route to uncertainty quantification. This is a property not reflected in point-estimate metrics such as MAE, RMSE, or $R^2$, but worth noting as a qualitative advantage even in cases when quantitative parity exists.

\begin{table}[!h]
\centering
\caption{Global intra-experiment reconstruction performance of the generative models, deterministic neural-network baselines, and regression baseline.}
\label{tab:intra_metrics}
\begin{tabular}{lccc}
\hline
\textbf{Method} & \textbf{MAE ($^\circ$C)} & \textbf{RMSE ($^\circ$C)} & \textbf{$R^2$} \\
\hline
GNN-GRU VAE        & 11.73 & 20.54 & 0.98 \\
GNN-GRU GAN        & 12.25 & 22.37 & 0.98 \\
GNN-GRU DM         & 13.23 & 22.61 & 0.97 \\
GNN-GRU            & 15.57 & 25.26 & 0.97 \\
GRU                & 19.26 & 35.22 & 0.96 \\
Regression         & 32.97 & 48.62 & 0.88 \\
\hline
\end{tabular}
\end{table}

\begin{figure}[!h]
\centering
\includegraphics[width=1\textwidth]{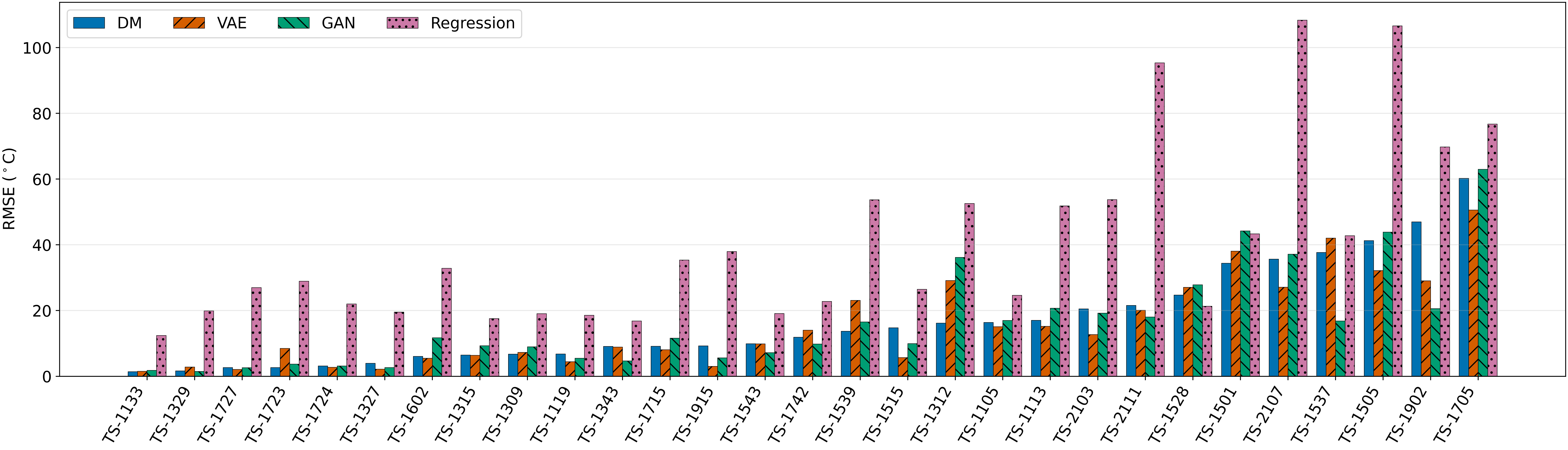}
\caption{Per-sensor intra-experiment RMSE comparison among the generative models and regression baseline.}
\label{fig:rmseWithin}
\end{figure}

The per-sensor RMSE comparison in Fig.~\ref{fig:rmseWithin} reveals substantial variation that is not apparent from the global averages. For sensors such as TS-1133, TS-1329, TS-1727, TS-1723, TS-1724, and TS-1327, the DM, VAE, and GAN achieve very low reconstruction errors, generally below $10^\circ$C, while ridge regression is consistently higher. The performance gap becomes much larger for sensors such as TS-1539, TS-1312, TS-1113, TS-2103, TS-2111, TS-2107, TS-1505, TS-1902, and TS-1705, where the regression RMSE rises to approximately $50^\circ$C--$110^\circ$C while the generative models remain mostly within the $15^\circ$C--$60^\circ$C range. Only a small number of sensors, including TS-1528 and TS-1501, show comparable performance between regression and the generative approaches. This pattern suggests that the global improvement of the generative models is distributed across much of the held-out sensor set rather than being driven by only a few favorable cases. The larger regression errors at the more difficult sensors may reflect regions in which the temperature response is strongly nonlinear and depends on broader spatial and temporal system behavior during the DCC transient. In contrast to ridge regression, which relies on a local linear combination of known measurements, the GNN--GRU generative architectures aggregate information across the sensor graph and integrate it over the temporal window, allowing them to represent longer-range and transient-dependent correlations more effectively.

\begin{figure}[!ht]
\centering
\includegraphics[width=0.95\textwidth]{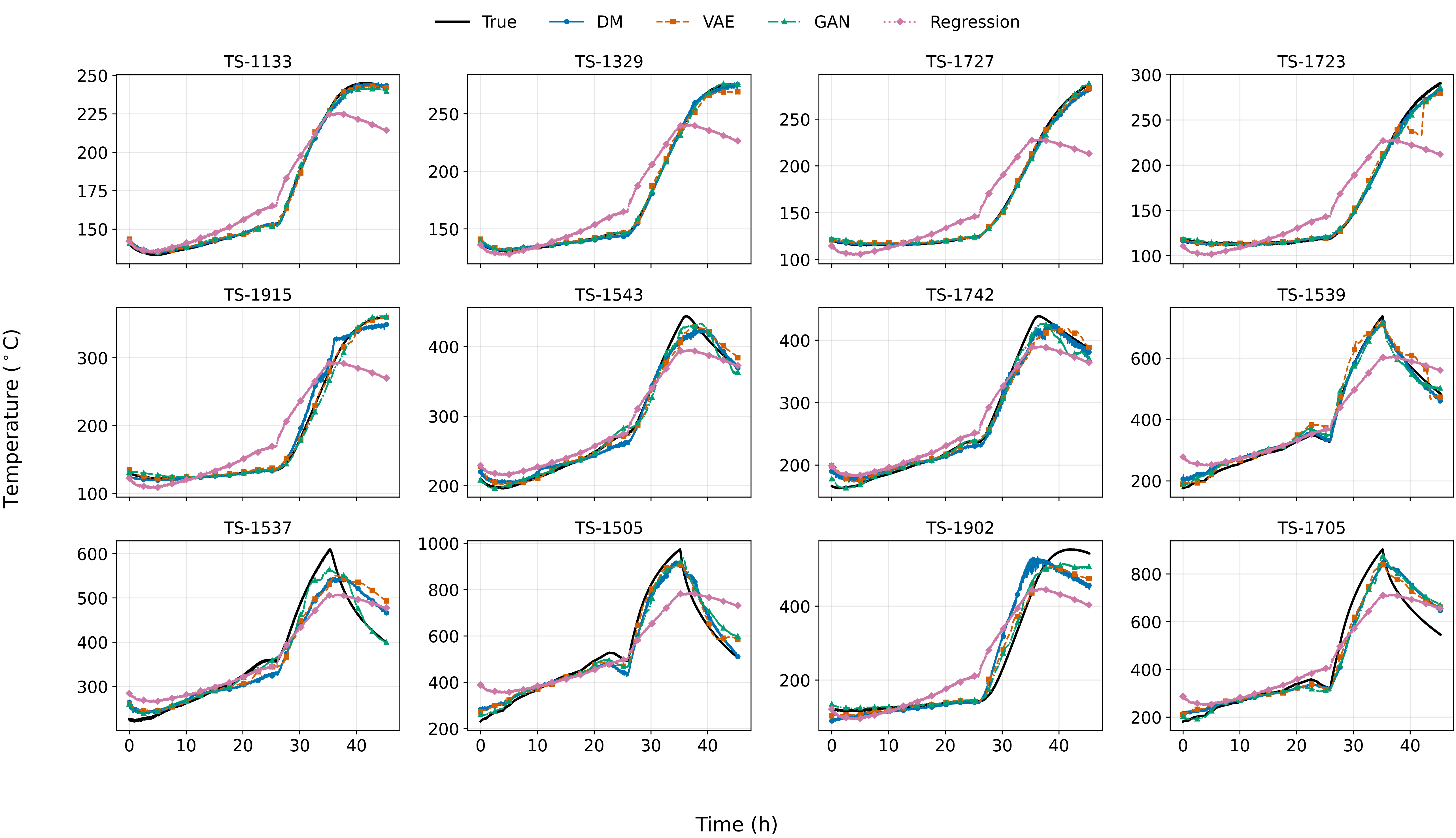}
\caption{Representative intra-experiment sensor reconstruction for selected fixed test sensors.}
\label{fig:withinTrend}
\end{figure}

The representative time-series traces in Fig.~\ref{fig:withinTrend} provide further insight into the sensor-wise error patterns. Across the $45.6$-hour PG{-}29 window, the three generative models track the true trajectory closely through both the quasi-steady early phase and the rapid heatup/natural-circulation ramp that occurred at around $25$--$40$~h. For the lower-error sensors, such as TS-1133, TS-1329, TS-1727, and TS-1723, the generative predictions remain close to the measured trajectories throughout the transient, including the final high-temperature plateau. For the more challenging sensors, including TS-1539, TS-1505, TS-1902, and TS-1705, the models still capture the timing and overall shape of the sharp heatup and subsequent peak, although larger differences appear in peak magnitude and late-time cooldown behavior. Ridge regression shows a different failure pattern: its response is excessively smooth, begins to deviate before the main temperature rise, and substantially underpredicts the peak or late-time temperature for several sensors, particularly TS-1133, TS-1329, TS-1727, TS-1723, TS-1915, TS-1505, TS-1902, and TS-1705. This behavior is consistent with the per-sensor RMSE results and indicates that the main advantage of the generative models lies not only in reducing average error, but also in preserving transient timing, nonlinear ramp behavior, and peak response across spatially diverse sensor locations. To maintain readability, Fig.~\ref{fig:withinTrend} presents representative low-, moderate-, and high-error cases from different HTTF regions; complete measured and reconstructed histories for all held-out thermocouples and reconstruction methods are provided in the supplementary material.

The generative models showed different computational tradeoffs. VAE and GAN required longer offline training, approximately 16 min and 3.1 h, respectively, but evaluation with the trained models was very fast, requiring only about 4 s for both. In contrast, DM required approximately 15 min for training and 12 min for evaluation because reconstruction relies on iterative reverse-diffusion sampling. The deterministic regression baseline remained substantially less computationally demanding. These results indicate that VAE and GAN are more immediately compatible with near-real-time digital-twin operation after offline training, whereas the current DM implementation would require inference-time optimization or reduced sampling before near-real-time deployment.

\begin{figure}[!h]
    \centering
    \includegraphics[width=0.95\linewidth]{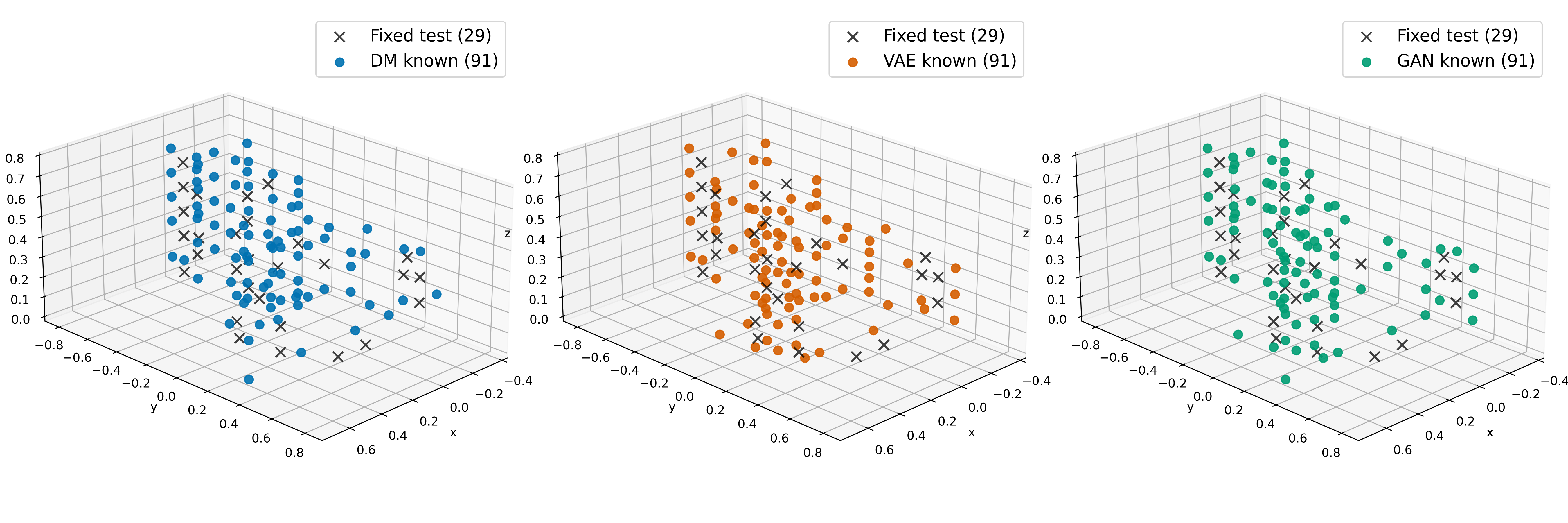}
    \caption{Three-dimensional locations of the known sensors used by DM, VAE, and GAN cases, overlaid with the fixed test sensors.}
    \label{fig:location}
\end{figure}

Fig.~\ref{fig:location} compares the spatial distribution of the known sensors used by the selected DM, VAE, and GAN cases. In all three cases, the known sensors are broadly distributed over the HTTF thermocouple domain and surround the fixed test sensors, rather than being concentrated in a single axial or radial region. This indicates that the comparison is not dominated by visibly different known-sensor placements among the three models. Therefore, the observed reconstruction differences are more likely associated with the reconstruction strategy and model formulation, despite still being affected by small differences in the specific known-sensor set.

\begin{figure}[!h]
    \centering
    \includegraphics[width=0.99\linewidth]{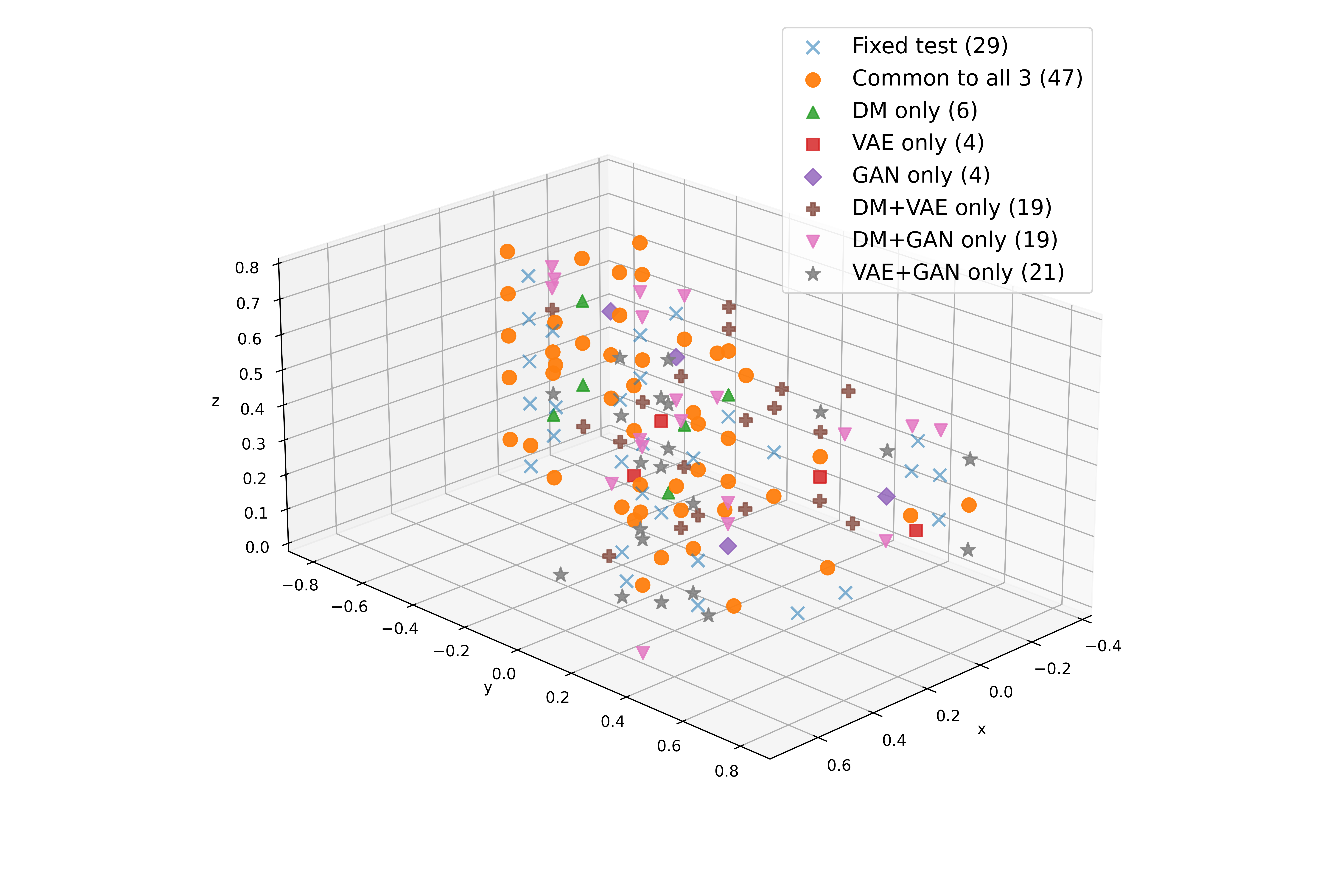}
    \caption{Spatial overlap of the known sensor sets used by the selected DM, VAE, and GAN cases.}
    \label{fig:overlap_location}
\end{figure}

Fig.~\ref{fig:overlap_location} further shows the overlap among the known-sensor sets, and the corresponding counts are summarized in Table~\ref{tab:known_sensor_overlap}. A large subset of 47 sensors is common to all three models, while only a small number of sensors are unique to each method: six for DM, four for VAE, and four for GAN. The remaining sensors are shared pairwise between two methods, with 19 sensors shared only by DM and VAE, 19 sensors shared only by DM and GAN, and 21 sensors shared only by VAE and GAN. Ultimately though, the substantial overlap reduces but does not altogether eliminate the possibility that differences in known-sensor selection contribute to the observed model-performance differences. At the same time, the model-specific and pairwise-only sensors may still influence local reconstruction behavior, especially for fixed test sensors located near sparsely covered regions.

\begin{table}[H]
\centering
\caption{Overlap summary of known sensor sets used by the selected DM, VAE, and GAN cases.}
\label{tab:known_sensor_overlap}
\begin{tabular}{lc}
\hline
\textbf{Sensor-set category} & \textbf{Number of sensors} \\
\hline
Fixed test sensors & 29 \\
Common to DM, VAE, and GAN & 47 \\
DM only & 6 \\
VAE only & 4 \\
GAN only & 4 \\
DM and VAE only & 19 \\
DM and GAN only & 19 \\
VAE and GAN only & 21 \\
\hline
\end{tabular}
\end{table}

\subsubsection{Effect of Reducing the Number of Known Sensors}
\label{subsec:known_sensor_reduction}

To evaluate the sensitivity of the reconstruction framework to increasingly sparse instrumentation, the number of known sensors was progressively reduced while retaining the same fixed test-sensor evaluation setting. Since the full comparison in the previous section includes all six models, this reduced-sensor study is reported only for the DM as a representative sensitivity case. Table~\ref{tab:known_sensor_reduction} summarizes the global reconstruction metrics for the original 91-known-sensor case and the reduced-known-sensor cases.

\begin{table}[H]
\centering
\caption{Effect of reducing the number of known sensors on intra-experiment DM reconstruction performance.}
\label{tab:known_sensor_reduction}
\begin{tabular}{lccc}
\hline
\textbf{Number of known sensors} & \textbf{MAE ($^\circ$C)} & \textbf{RMSE ($^\circ$C)} & \textbf{$R^2$} \\
\hline
91 & 13.23 & 22.61 & 0.97 \\
77 & 16.46 & 27.40 & 0.96 \\
61 & 17.24 & 29.62 & 0.95 \\
45 & 17.93 & 31.05 & 0.94 \\
\hline
\end{tabular}
\end{table}

As expected, reconstruction performance gradually decreases as fewer known sensors are available for conditioning. Reducing the known-sensor count from 91 to 77 increases the RMSE from 22.61$^\circ$C to 27.40$^\circ$C, and further reductions to 61 and 45 known sensors increase the RMSE to 29.62$^\circ$C and 31.05$^\circ$C, respectively. The corresponding $R^2$ values remain high, decreasing from 0.97 for 91 known sensors to 0.94 for 45 known sensors. This indicates that the DM retains strong reconstruction capability even when the conditioning information is substantially reduced.

\subsection{Inter-Experiment Reconstruction}
\label{sec:inter}

The inter-experiment protocol, in which the models are trained on PG{-}27 (PCC) and evaluated on distinct PG{-}29 (DCC), presents a substantially more difficult generalization problem. As shown in Table~\ref{tab:cross_global_metrics}, the DM achieves the best transfer performance, with an MAE of $30.57\,^\circ$C, an RMSE of $42.84\,^\circ$C, and an $R^2$ of $0.91$. It is followed by the VAE, GAN, deterministic GRU, and regression models. All three generative models outperform regression in MAE, RMSE, and $R^2$, while the deterministic GRU also provides an improvement over regression. Overall, the results demonstrate that incorporating target-specific physical and transient context improves cross-experiment reconstruction. However, the higher errors relative to the intra-experiment results highlight the difficulty of transferring learned spatiotemporal relationships between PCC and DCC transients conducted in the same experimental facility.

\begin{table}[!ht]
\centering
\caption{Global inter-experiment reconstruction performance of the generative models, deterministic GRU baselines, and regression baseline.}
\label{tab:cross_global_metrics}
\begin{tabular}{lccc}
\hline
\textbf{Method}  & \textbf{MAE ($^\circ$C)} & \textbf{RMSE ($^\circ$C)} & \textbf{$R^2$} \\
\hline
section-context DM         & 30.57 & 42.84 & 0.91 \\
section-context VAE        & 31.03 & 49.67 & 0.88 \\
section-context GAN        & 35.85 & 50.82 & 0.87 \\
section-context GRU        & 36.83 & 54.21 & 0.86 \\
GRU        & 55.91 & 79.34 & 0.70 \\
Regression & 38.52 & 64.25 & 0.80 \\

\hline
\end{tabular}
\end{table}

A noteworthy shift relative to intra-experiment reconstruction is that the relative ordering among the three generative models changes: intra-experiment, VAE $>$ GAN $>$ DM (by a narrow margin); inter-experiment, DM $>$ VAE $>$ GAN. The DM's relative standing improves specifically under distribution shift, consistent with the additional temperature-space, local-neighbor, section-context, transient-slope, and residual-magnitude loss terms built into its inter-experiment (section-context) formulation, all of which are designed to regularize exactly the kind of transient-shape and spatial-context transfer that PCC-to-DCC generalization requires. This suggests that the diffusion framework's principal benefit is realized less in raw interpolation accuracy under matched conditions and more in its capacity to be shaped, via its conditioning and loss structure, toward robust generalization---an appealing property for a generative digital twin that must eventually operate across transients not seen during training.

\begin{figure}[!ht]
\centering
\includegraphics[width=0.95\textwidth]{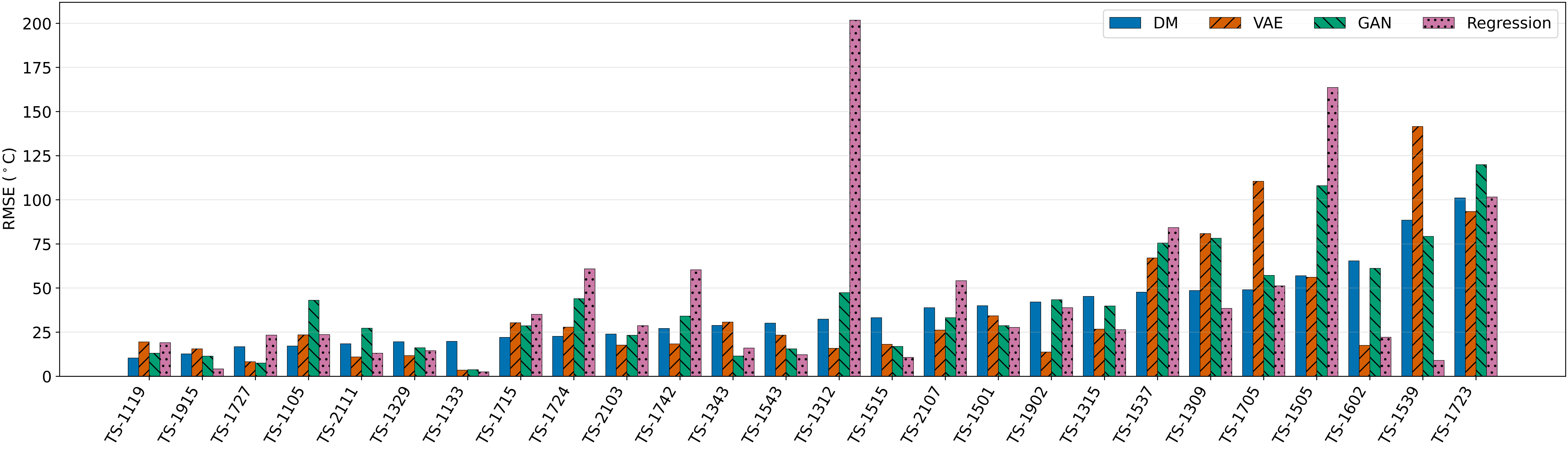}
\caption{Per-sensor inter-experiment RMSE comparison for PG{-}27 PCC to PG{-}29 DCC transfer.}
\label{fig:rmseCross}
\end{figure}

\begin{figure}[!ht]
\centering
\includegraphics[width=0.95\textwidth]{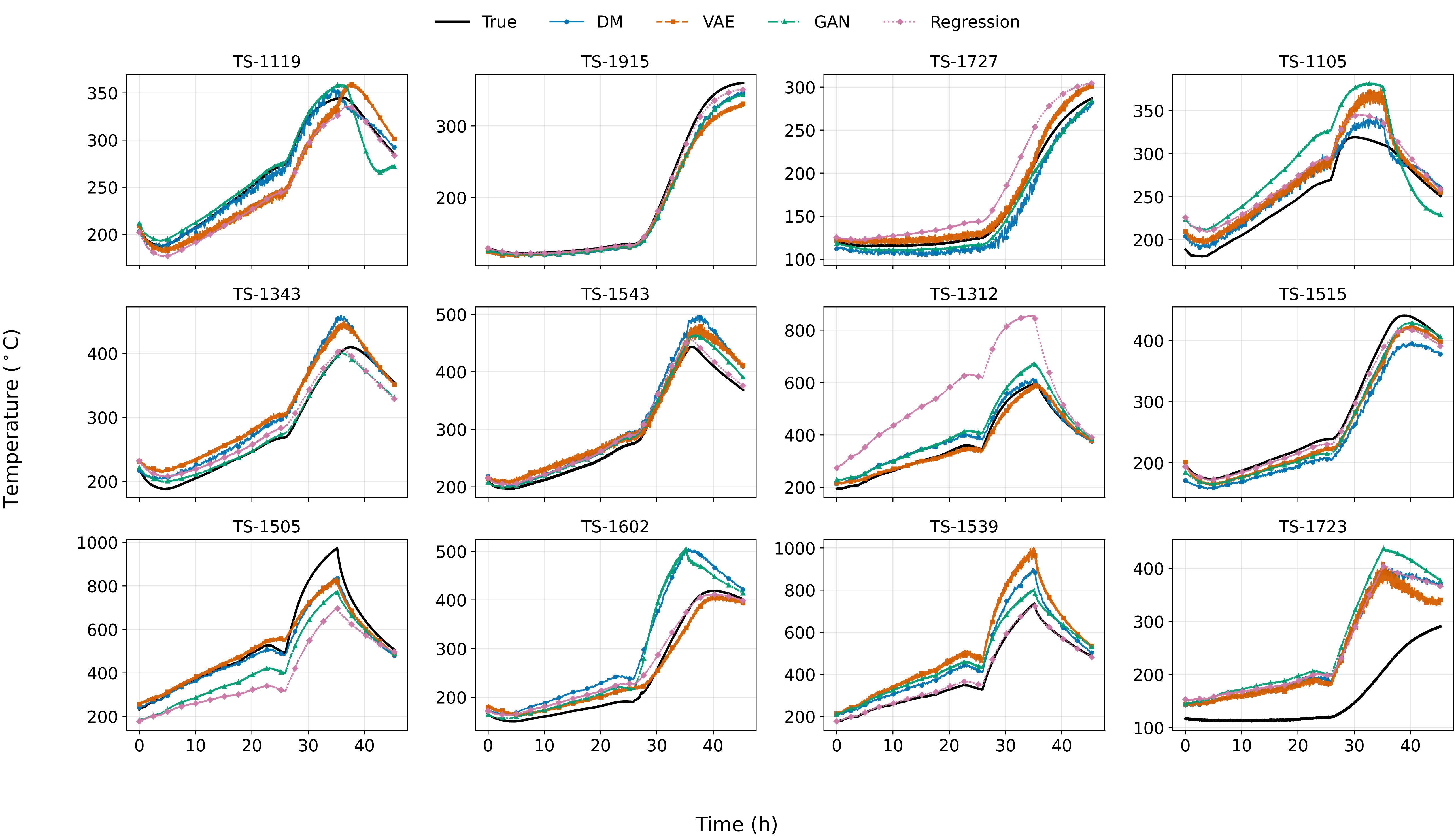}
\caption{Representative inter-experiment reconstruction from PCC-to-DCC transfer.}
\label{fig:crossTrend}
\end{figure}

The transition from intra- to inter-experiment evaluation increases the RMSE for every method, although the proportional degradation differs substantially. The DM RMSE increases from $22.61^\circ$C to $42.84^\circ$C, while the VAE and GAN RMSE values increase from $20.54^\circ$C and $22.37^\circ$C to $49.67^\circ$C and $50.82^\circ$C, respectively. The deterministic GRU and regression baselines increase from $35.22^\circ$C and $48.62^\circ$C to $54.21^\circ$C and $64.25^\circ$C, respectively. This is consistent with the physical differences between the two transients, as summarized in Table ~\ref{tab:pg27_pg29_key_features}: PG{-}27 is a PCC driven by loss of forced convection under an intact, pressurized boundary, whereas PG{-}29 is a DCC initiated by a crossover-duct break, introducing early-time helium--nitrogen gas exchange, lock-exchange stratification, and a subsequent natural-circulation regime with no analogue in the PCC training data. A model trained exclusively on PCC dynamics must therefore extrapolate to physical regimes it never observed---a fundamentally harder task than interpolating within a single, internally consistent transient---thus explaining why every model's error increases substantially even as the relative advantage of the best generative model over the baseline is preserved (the DM shows a $33\%$ RMSE reduction over regression in the inter-experiment case, comparable to the relative gain observed in the intra-experiment case).

The per-sensor RMSE comparison (Fig.~\ref{fig:rmseCross}) shows this degradation to be highly non-uniform. Several sensors (TS-1105, TS-1312, TS-1505) exhibit large errors for at least one model---most visibly the GAN, which spikes well above the other two generative models on a handful of channels---while regression shows an extreme outlier at TS-1312 ($\sim 200^\circ$C RMSE), the largest error observed anywhere in the study. These sensors plausibly correspond to locations most sensitive to DCC-specific phenomena (e.g., the lower-plenum or transition zones affected by natural circulation onset, or the centerline higher-temperature region), for which the PCC training data provide the least representative prior. Even so, the DM remains among the lowest-error methods for nearly every sensor in Fig.~\ref{fig:rmseCross}, underscoring its comparative robustness under transfer.

One sensor, TS-1723, merits separate discussion (Fig.~\ref{fig:crossTrend}) for standing out as qualitatively different from the rest of the per-sensor comparison. For nearly every other sensor, the four methods disagree with each other in magnitude, with at least one tracking the true response reasonably closely. For TS-1723, by contrast, all four methods---DM, VAE, GAN, and regression---converge onto a similar, incorrect trajectory, plateauing well above the true response for most of the transient rather than spreading out around it. All three generative models reconstruct this sensor accurately under the intra-experiment protocol (Fig.~\ref{fig:withinTrend}), but this does not resolve whether TS-1723 reflects a genuine, learnable physical coupling to its neighborhood or is itself a channel reconstructed via backup-sensor interpolation (Section~\ref{sec:predata}). Consequently, the TS{-}1723 result should be interpreted with caution since the comparison is against an interpolated reference rather than an independent measured thermocouple signal.

\subsection{Discussion and Implications}

\label{sec:discussion}

Taken together, Sections~\ref{sec:intra} and \ref{sec:inter} demonstrate that, within a single transient, generative reconstruction can achieve high accuracy from sparse sensor measurements. All three generative architectures reduce reconstruction error by roughly half relative to the deterministic regression baseline while achieving $R^2 > 0.97$. These results indicate that the generative models can effectively recover the spatial and temporal relationships needed to reconstruct unobserved sensor responses under nominal, in-distribution conditions. Such performance supports the potential use of sparse-sensor reconstruction within digital twins, particularly for near-real-time state estimation and sensor-criticality analysis when the operating conditions are sufficiently represented by the training data.

The inter-experiment results provide a more demanding assessment of this capability and show that the generative advantage persists under domain shift, although with substantially increased reconstruction error. All three generative models continue to outperform regression when transferring from the PG{-}27 PCC transient to the PG{-}29 DCC transient, with the DM showing the strongest relative performance under this more challenging setting. This improvement is consistent with the temperature-space, local-neighbor, section-context, transient-slope, and residual-magnitude loss terms incorporated into its inter-experiment formulation to promote transient-shape and spatial-context transfer. Nevertheless, the absolute reconstruction error increases substantially for all methods relative to the intra-experiment case, demonstrating that generalization across distinct operating regimes remains a major challenge. These findings motivate training on a broader range of transients so that the conditioning distribution better represents expected off-normal behavior, as well as incorporating physics-informed constraints, such as conservation-based penalties, to regularize extrapolation beyond the training distribution.

The sensor-level analysis also demonstrates that large reconstruction errors should not automatically be interpreted as model failures. The TS{-}1723 case in Section~\ref{sec:inter} is particularly informative because all methods, including the structurally distinct regression baseline, deviate from the reference response in a similar direction. In this dataset, such behavior may instead reflect limitations in the reference signal associated with the backup-channel interpolation used to replace non-functioning primary thermocouple measurements. This observation has broader methodological implications for sparse-sensor reconstruction using legacy experimental datasets: large per-sensor errors should be examined against the instrumentation record and data-processing history before being attributed solely to deficiencies in the reconstruction model. Otherwise, limitations in the reference data may understate actual model performance or lead to misleading comparisons among reconstruction architectures.

Finally, the present framework extends earlier HTTF virtual-sensing studies by addressing two important limitations related to evaluation and inference. Previous work demonstrated that long-short-term-memory- and GRU-based models can reconstruct sensor responses from known measurements and that variational Bayesian last-layer augmentation can provide uncertainty-aware predictions with high accuracy and limited additional computational cost~\cite{farha2025sensor,burnett2026variational}. However, these studies were evaluated primarily within a single experimental dataset and relied on one-step-ahead forecasting, in which ground-truth measurements from preceding time steps were supplied during prediction. These assumptions provide limited evidence of generalization across different experimental conditions and may be impractical for autonomous deployment when measurements at the reconstructed locations are unavailable. The present study addresses these limitations by evaluating reconstruction across distinct HTTF experiments and eliminating dependence on ground-truth sequential inputs during inference, thereby providing a more stringent assessment of sparse-sensor reconstruction under conditions relevant to practical deployment.

\section{Conclusions}
\label{sec:conc}

This study evaluated six methods for reconstructing withheld HTTF TS thermocouple measurements under intra- and inter-experiment protocols. Under the PG{-}29 intra-experiment protocol, the three GNN--GRU generative models outperformed the deterministic GNN--GRU, plain GRU, and ridge-regression baselines. The GNN--GRU VAE achieved the lowest aggregate error, with an MAE of $11.73\,^\circ$C, an RMSE of $20.54\,^\circ$C, and an $R^2$ of 0.98. The GNN--GRU GAN and DM achieved RMSE values of $22.37\,^\circ$C and $22.61\,^\circ$C, respectively. Relative to ridge regression, the three generative models reduced the RMSE by approximately 53\%--58\%. The deterministic GNN--GRU reduced the RMSE by approximately 28\% relative to the plain GRU, demonstrating the value of combining spatial and temporal information. The known-sensor analysis showed substantial, but incomplete, overlap among the selected DM, VAE, and GAN sensor sets. In the diffusion-model sensitivity study, reducing the number of known sensors from 91 to 45 increased the RMSE from $22.61\,^\circ$C to $31.05\,^\circ$C, while the $R^2$ decreased from 0.97 to 0.94. The inter-experiment protocol provided a more demanding evaluation by training on PG{-}27 PCC data and testing on the physically distinct PG{-}29 DCC transient. The section-context DM achieved the best transfer performance, with an MAE of $30.57\,^\circ$C, an RMSE of $42.84\,^\circ$C, and an $R^2$ of 0.91. It reduced the RMSE by approximately 33\% relative to ridge regression. The section-context GRU also substantially outperformed the plain GRU, demonstrating the value of target-specific physical and transient-context information under cross-experiment transfer.

Overall, the results support spatial--temporal generative modeling as a promising approach for reconstructing missing thermocouple measurements in safety-critical digital twin applications. The larger and spatially nonuniform errors observed under cross-experiment transfer indicate that generalization across physically different transients remains the principal challenge. Future work should evaluate additional HTTF transients, perform controlled ablations of the section-context and composite-loss terms, assess predictive-uncertainty calibration, and examine robustness to measurement noise, sensor drift, intermittent failure, and dynamically changing sensor availability.

\section*{Data Availability}
\label{sec:avail}

The authors have created a private GitHub repository containing all the datasets and codes necessary to reproduce all the results of this work. To ensure confidentiality of this research, the authors will make this repository public during an advanced stage of the review process, and it will be listed under their research group's public Github page: \url{https://github.com/aims-umich}.

\section*{Acknowledgment}

This work was supported through Idaho National Laboratory (INL) Laboratory Directed Research and Development (LDRD) Program Award 24A1081-116FP, under U.S.~Department of Energy Idaho Operations Office Contract DE-AC07-05ID14517.

\section*{CRediT Author Statement}

\begin{itemize}
    \item \textbf{Umme Mahbuba Nabila}: Conceptualization, Methodology, Software, Validation, Formal Analysis, Visualization, Investigation, Data Curation, Writing - Original Draft. 
    \item \textbf{Linyu Lin}: Conceptualization, Methodology, Resources, Funding Acquisition, Supervision, Project Administration, Writing - Original Draft.
    \item \textbf{Majdi I. Radaideh}: Conceptualization, Methodology, Resources, Funding Acquisition, Supervision, Project Administration, Writing - Original Draft. 
\end{itemize}

%\section*{References}

\bibliographystyle{elsarticle-num}
%\bibliographystyle{apa}
%\biboptions{authoryear}
\setlength{\bibsep}{0pt plus 0.3ex}
{
\bibliography{references}}

\end{document}